\documentclass[conference,a4paper]{IEEEtran}
\IEEEoverridecommandlockouts

\usepackage[hidelinks]{hyperref}
\usepackage[cmex10]{amsmath}
\usepackage{amssymb,amsfonts}
\usepackage{dblfloatfix}

\usepackage[ruled,vlined]{algorithm2e}
\usepackage{graphicx}
\graphicspath{{Figures/PDF/}{Figures/PNG/}}

\usepackage{booktabs}
\usepackage{siunitx}
\usepackage[numbers,compress]{natbib}
\usepackage{texnames}
\usepackage{bm,bbm}
\usepackage{orcidlink}
\usepackage{multirow}

\begin{document}


\title{\uppercase{Low-Rank Adaptation of Geospatial Foundation Models for Wildfire Mapping Using Sentinel-2 Data}

\thanks{This research is part of the EO-AI4GlobalChange project funded by Digital Futures, Stockholm, Sweden.
}}

\author{	\IEEEauthorblockN{Ali Shibli\orcidlink{0009-0001-0794-6443}}
	\IEEEauthorblockA{\textit{KTH Royal Institute of Technology}\\
		Stockholm, Sweden\\
		shibli@kth.se}
	\and
	\IEEEauthorblockN{Andrea Nascetti\orcidlink{0000-0001-9692-8636}}
	\IEEEauthorblockA{\textit{KTH Royal Institute of Technology}\\
		Stockholm, Sweden\\
		nascetti@kth.se}	\and
	\IEEEauthorblockN{Yifang Ban\orcidlink{0000-0003-1369-3216}}
	\IEEEauthorblockA{\textit{KTH Royal Institute of Technology}\\
		Stockholm, Sweden\\
		yifang@kth.se}
}

\maketitle
\begin{abstract}

Wildfire burned-area mapping is essential for damage assessment, emissions modeling, and understanding fire–climate interactions across diverse ecological regions. Recent geospatial foundation models provide strong general-purpose representations for satellite imagery, yet there is still no clear understanding of how to efficiently adapt these models for downstream Earth observation tasks, particularly under geographic and temporal domain shift. This study evaluates three state-of-the-art Geospatial Foundation Models (GFMs) - Terramind, DINOv3, and Prithvi-v2 - for burned-area mapping across the United States and Canada using Sentinel-2 data. Leveraging 3,820 wildfire events from 2017–2023, we conduct spatial and temporal generalization tests across diverse biomes. We systematically compare full fine-tuning, decoder-only fine-tuning, and Low-Rank Adaptation (LoRA) for adapting each model. Across all experiments, LoRA provides the strongest cross-domain generalization while updating less than 1\% of parameters, demonstrating a favorable trade-off between accuracy and efficiency. Prithvi-v2 with LoRA achieves the highest overall accuracy and the larger improvement compare to full fine-tuning. These findings indicate that geospatial foundation models, when adapted using lightweight parameter-efficient methods such as LoRA, offer a robust and scalable solution for large-scale burned-area mapping. Code is available at \url{https://github.com/alishibli97/wildfire-lora-gfm}.

\end{abstract}


\section{Introduction}

Wildfires are a major driver of landscape change and greenhouse gas emissions. Accurate Burned-Area (BA) mapping includes post-fire impact assessment, carbon and aerosol emission estimates, and the evaluation of fire–climate feedbacks at regional to global scales. 
Freely accessible high-resolution optical missions such as Sentinel-2 and Landsat have enabled burned-area mapping at 10–30 m resolution. Early approaches relied on spectral indices such as the Normalized Burn Ratio (NBR) and its temporal difference, dNBR, often combined with thresholding or Object-Based Image Analysis (OBIA) to delineate burned patches \cite{al2022burned,suwanprasit2024mapping}. These methods improve spatial detail but still struggle with variable illumination, partial burning, mixed pixels, and confusion with other disturbance types, and fixed thresholds or hand-designed rules can be difficult to transfer across regions or vegetation types.

Deep learning has become the standard for high-resolution BA mapping in recent years. U-Net and related encoder–decoder architectures have been widely applied to mono-temporal and bi-temporal Sentinel-2 imagery, yielding significant accuracy gains over index-based methods \cite{knopp2020deep, brand2021semantic}. 
UNet variants that incorporate attention modules, and Siamese or bi-temporal networks that better leverage scene change information. For example, BiAU-Net introduces bi-temporal attention and a task-specific loss to better capture fine-scale burned edges and small patches, and evaluates performance across diverse regions on multiple continents \cite{sui2024biau}. 
Despite advances in deep-learning–based burned-area mapping, most studies remain geographically or temporally narrow: they focus on a single country, biome, or fire season. Some studies attempt transfer learning or domain adaptation when shifting to different fire regimes or land-cover types, but these typically deal with small agricultural burns or limited cross-region scenarios \cite{anand2025domain}. 

In parallel, the remote-sensing community has begun to adopt geospatial foundation models  (GFMs) such as Prithvi-EO \cite{prithvi}, TerraMind \cite{terramind}, Dinov3 \cite{simeoni2025dinov3} that are pre-trained on massive multispectral and multi-temporal satellite imagery collections and fine-tuned for various tasks. These GFMs have shown strong performance on segmentation and change-detection benchmarks \cite{marsocci2024pangaea, simumba2025geo}, but their application to wildfire burned-area mapping at large scale remains underexplored. 
However, adapting such large models to specific EO tasks presents practical challenges. Full fine-tuning of hundreds of millions of parameters is computationally expensive and difficult to maintain for multiple regions or time periods. Decoder-only fine-tuning offers a cheaper alternative but may fail to capture domain-specific variations in the encoder, reducing its robustness under domain shift. Parameter-efficient fine-tuning (PEFT) methods address this by updating only a small subset of weights or injecting trainable low-rank modules into frozen backbones \cite{han2024parameter, xin2024parameter}. A recent work has successfully applied PEFT to geospatial foundation models showing that PEFT can match or exceed full fine-tuning performance on multiple Earth-observation tasks while reducing training cost and improving generalization to new geographic regions \cite{marti2025fine}. Low-Rank Adaptation (LoRA) is one such method: it introduces trainable low-rank matrices into linear layers while keeping the original weights frozen, greatly reducing task-specific parameters while often matching or surpassing full fine-tuning performance \cite{hu2022lora}. Yet, it remains unclear which adaptation strategy — full fine-tuning, decoder-only fine-tuning, or PEFT — offers the best trade-off for large-scale burned-area mapping.


In this research, we address this gap by evaluating three state-of-the-art geospatial foundation models - TerraMind, DINOv3, and Prithvi-v2 - for wildfire burned-area mapping across the United States and Canada using Sentinel-2 imagery. We systematically compare full fine-tuning, decoder-only fine-tuning, and Low-Rank Adaptation (LoRA) to assess how different adaptation strategies affect performance and generalization under geographic and temporal domain shift. 



\section{Method}

\begin{figure}
    \centering
    \includegraphics[width=0.9\linewidth]{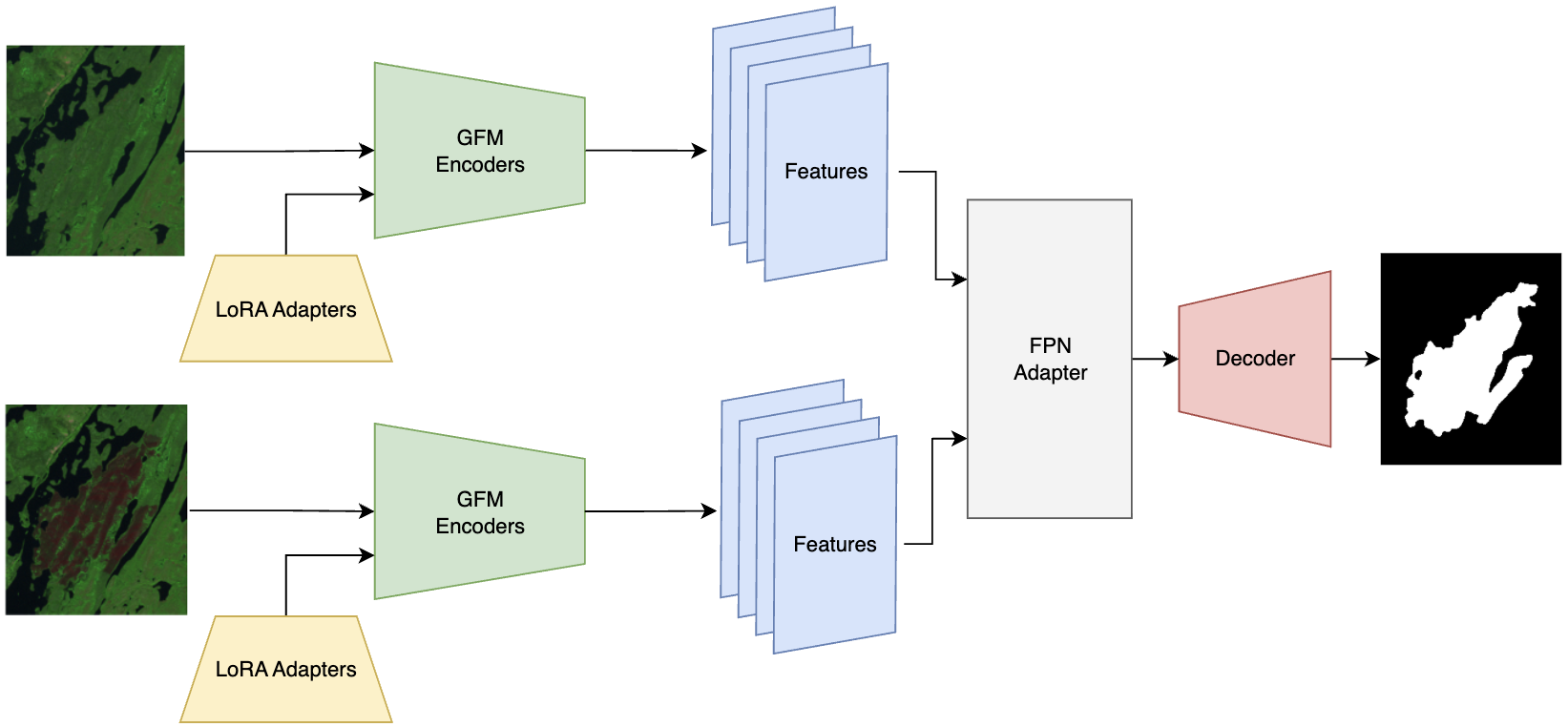}
    \caption{Overview of the proposed method. Bi-temporal images (pre- and post- wildfire) are passed separately through the same GFM encoder with LoRA Adapter applied to attention modules. Features are then extracted and combined via FPN Adapter, and finally decoded to predict the burned area.}
    \label{fig:overview}
\end{figure}

We treat burned area mapping as a change detection problem. Given a pre-fire image $x^{\mathrm{pre}}$ and a post-fire image $x^{\mathrm{post}}$, our aim is to train a model that predicts a binary change mask $y \in \{0,1\}^{H \times W}$ indicating burned vs.\ unburned pixels. Our overall architecture consists of 
(i) a shared Transformer-based encoder with optional LoRA adapters, 
(ii) a learned pyramidal neck that converts encoder features into multi-scale feature maps, 
(iii) bi-temporal change fusion via concatenation of pre- and post-fire features, 
and (iv) a UPerNet decoder producing dense predictions. An overview of our proposed method is illustrated in Fig.~\ref{fig:overview}.

\paragraph{Problem Formulation}

Let $x^{\mathrm{pre}}, x^{\mathrm{post}} \in \mathbb{R}^{C \times H \times W}$ denote pre- and post-fire Sentinel-2 reflectance patches. The model learns a function
\begin{equation}
    f_\theta : (x^{\mathrm{pre}}, x^{\mathrm{post}}) \rightarrow \hat{y},
\end{equation}
where $\hat{y} \in \mathbb{R}^{2 \times H \times W}$ are per-pixel logits for burned and unburned classes. We share a single backbone encoder for the two times and keep its weights frozen; the trainable parameters $\theta$ are the LoRA adapters (when enabled), the pyramidal neck, the decoder, and the final classification head.

\paragraph{LoRA Adaptation} 

We evaluate three GFM backbones: Prithvi-v2 \cite{prithvi}, TerraMind \cite{terramind}, and DINOv3 \cite{simeoni2025dinov3}. Each backbone is a Vision Transformer encoder that maps an input image to a sequence of tokens. 

To enable parameter-efficient fine-tuning, we insert Low-Rank Adaptation (LoRA) into selected projection layers. Given a weight matrix $W \in \mathbb{R}^{d_{\text{out}} \times d_{\text{in}}}$, LoRA augments it as
\begin{equation}
    W' = W + \Delta W, \qquad 
    \Delta W = B A,
\end{equation}
where $A \in \mathbb{R}^{r \times d_{\text{in}}}$ and $B \in \mathbb{R}^{d_{\text{out}} \times r}$ are low-rank trainable matrices with $r \ll d_{\text{in}}$, while the original weights $W$ are frozen. The adapted layer computes $y = W x + \alpha \Delta W x$ with a scaling factor $\alpha$.

In all three backbones we apply LoRA to the self-attention and projection layers in every Transformer block. For Prithvi-v2 we additionally insert LoRA adapters into the MLP layers and the 3D patch-embedding convolution via $1\times 1$ convolutional adapters. For all models, we freeze the entire encoder and only the LoRA adapters are trainable. 

\paragraph{From Tokens to Multi-Scale Feature Maps}

For each backbone we extract features from a small set of selected Transformer blocks $\ell \in \mathcal{L}$.
In the case of Prithvi-v2, we return 2D feature maps from the chosen blocks. For TerraMind and DINOv3, we convert the token sequences $\mathbf{T}_{\ell} \in \mathbb{R}^{N \times D}$ into spatial maps by dropping the class token (when present), reshaping the remaining tokens to a $\sqrt{N} \times \sqrt{N}$ grid, and permuting dimensions to obtain
\begin{equation}
    \mathbf{F}_{\ell} \in \mathbb{R}^{D \times h_\ell \times w_\ell}.
\end{equation}

These per-layer feature maps are passed through a learned pyramidal neck, that plays the role of an FPN-like feature pyramid, which interpolates and projects them into four multi-scale feature levels:
\begin{equation}
    \mathcal{P}^{\mathrm{pre}} = \{P^{\mathrm{pre}}_1,\ldots,P^{\mathrm{pre}}_4\}, \qquad
    \mathcal{P}^{\mathrm{post}} = \{P^{\mathrm{post}}_1,\ldots,P^{\mathrm{post}}_4\}.
\end{equation}

At each pyramid level $k$, we fuse the two streams by channel-wise concatenation 
\( Z_k = \mathrm{concat}\!\left(P^{\mathrm{pre}}_k, \, P^{\mathrm{post}}_k\right)\),
yielding a bi-temporal pyramid 
\(
    \mathcal{Z} = \{Z_1, Z_2, Z_3, Z_4\}.
\)
This fusion mechanism is implemented identically in all three encoder--decoder variants.

\paragraph{Decoder and Mask Prediction}

The merged feature pyramid $\mathcal{Z}$ is fed into a UPerNet decoder \cite{xiao2018unified}. The decoder aggregates information across scales and produces a dense feature map, which a final $1\times 1$ convolution maps to 2 logits (burned vs.\ unburned). Bilinear interpolation upsamples the logits to the original patch size $(H,W)$:
\begin{equation}
    \hat{y} = \mathrm{upsample}\big(\mathrm{Conv}_{1\times 1}(\mathrm{UPerNet}(\mathcal{Z}))\big) \in \mathbb{R}^{2 \times H \times W},
\end{equation}
followed by a pixelwise softmax.

\paragraph{Loss Function}

Because burned pixels are typically a minority class, we apply class-balanced cross-entropy:
\begin{equation}
    \mathcal{L} = -\sum_{i,j} 
    w_{y_{ij}} \log p\big(\hat{y}_{ij} = y_{ij}\big),
\end{equation}
with $w_{\text{burn}} : w_{\text{unburn}} = 3 : 1$.

\section{Study Area and Data}


Our study covers 3,800 wildfire events across the United States and Canada between 2017 and 2023. Events were selected from the MTBS (US) \cite{mtbs} and NBAC (Canada) \cite{nbac} burned-area inventories, retaining only fires larger than 100 ha to ensure reliable spatial extent and reduce label noise. Sentinel-2 imagery (B4, B8, B12; 10--20 m) was filtered by cloud ($\leq 20\%$), snow ($\leq 20\%$), and missing-data coverage ($\leq 20\%$) over each event region. Burned-area labels were obtained by rasterizing official MTBS and NBAC fire perimeter polygons onto the Sentinel-2 grid. To evaluate generalization under temporal domain shift, we split the data into: 
\begin{equation}
    \mathcal{D}_{\text{source}} = \{\text{fires in 2017--2020}\}, 
    \mathcal{D}_{\text{target}} = \{\text{fires in 2021--2023}\}.
\end{equation}

\begin{figure}
    \centering
    \includegraphics[width=0.8\linewidth]{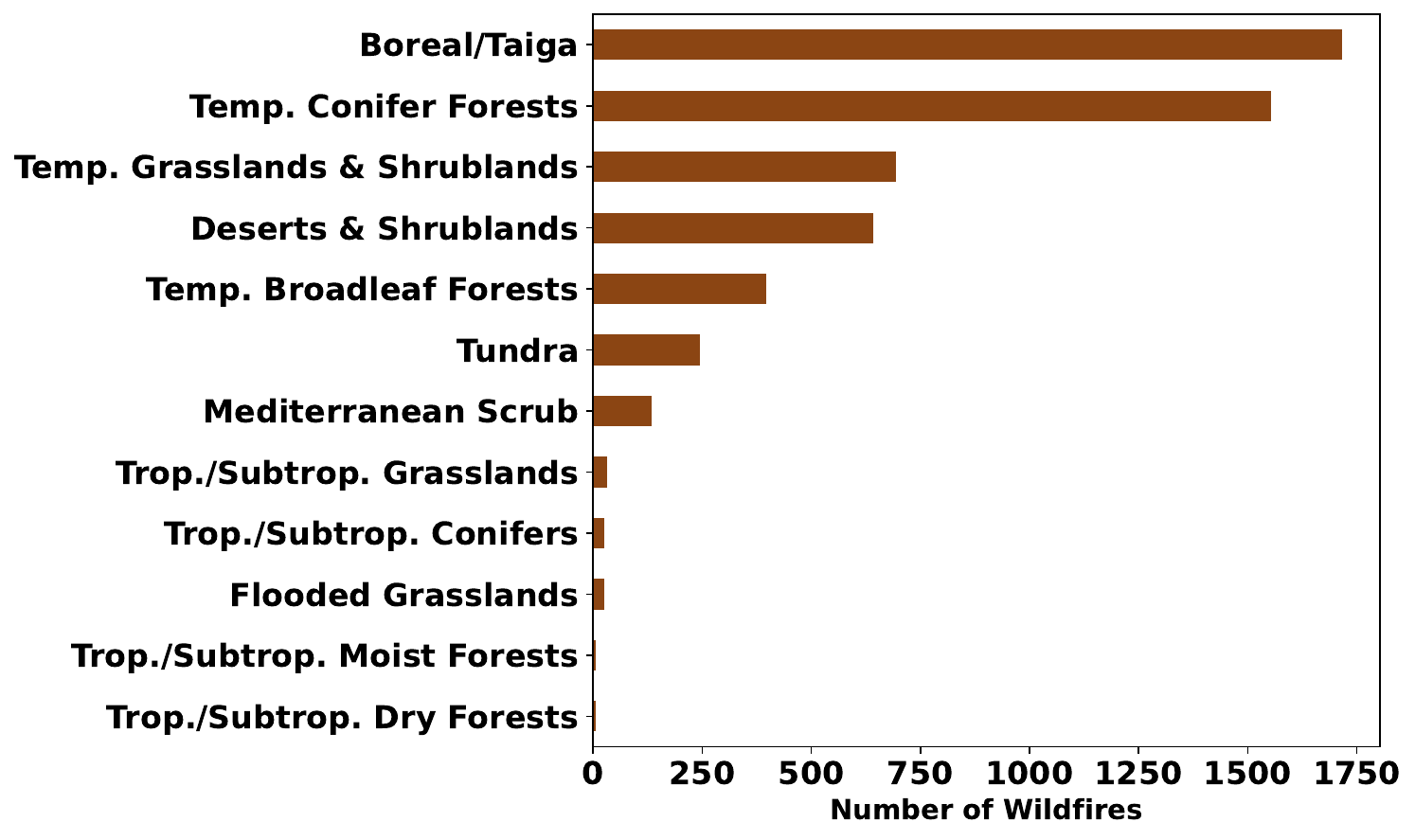}
    
    \caption{Distribution of wildfire events per biome in the US and Canada (2017-2023)}
    \label{fig:biome_distribution}
\end{figure}

\begin{figure}
    \centering
    \includegraphics[width=0.8\linewidth]{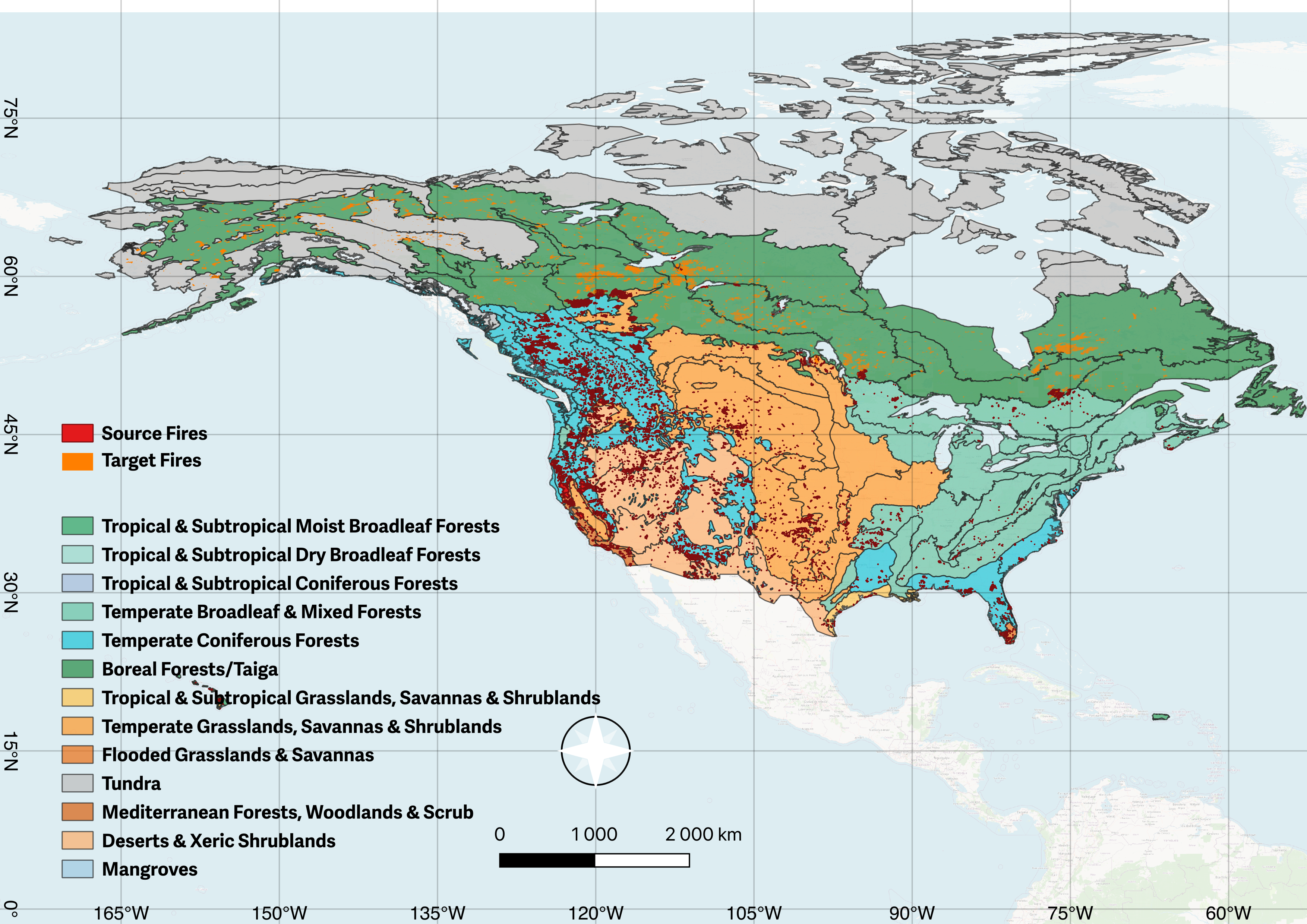}
    \caption{Spatiotemporal split of fires across the US and Canada}
    \label{fig:sp}
\end{figure}

Wildfire events are further split by terrestrial biomes to stress on ecological robustness. The biome distributions in our dataset are illustrated in Fig.~\ref{fig:biome_distribution}. This setting simulates real-world operational deployment where models must generalize to unseen fires across later years and different ecological regions. We take fires within Boreal/Taiga and Tundra biome regions as target domain fires, since they are sensitive to climate changes in fire regimes. For the source domain, we take fires in all the rest of the biomes. This results in $2298$ fires for training and $1522$ fires for testing. Fig.~\ref{fig:sp} illustrates all the biomes and the source and target fires on the map. All images are at 10meter resolutions and patched to  size $128 \times 128$. The ground true fire masks are derived from official fire perimeters: MTBS database in the US and NBAC in Canada.

\section{Experiments}

We evaluate three state-of-the-art GFMs: TerraMind (transformer-based multimodal EO model), DINOv3 (self-supervised vision transformer), and Prithvi-v2 (ViT-based EO foundation model). For each backbone, we train three variants: (i) full fine-tuning, (ii) decoder-only fine-tuning, and (iii) LoRA-adapted model. All models use the same lightweight UPerNet decoder and are trained with the same objective. Table~\ref{tab:params_lora} summarizes the parameter counts for each backbone before and after LoRA adaptation. When LoRA is applied to the encoder, TerraMind and DINOv3 require approximately 0.51\% trainable parameters and Prithvi-v2 requires 1.03\%. When the full network (encoder--decoder) is considered, the proportion of trainable parameters is 15.25\% for TerraMind, 15.23\% for DINOv3, and 6.79\% for Prithvi-v2. Across all models, LoRA provides a substantial reduction in the number of trainable parameters compared to full fine-tuning.

We train all models using Adam with a learning rate of $1\times10^{-4}$, a batch size of 2, and best model based on validation IoU. LoRA adapters use rank $r=8$ and scaling $\alpha=1.0$, and are applied to the attention
projections of each backbone while keeping all original encoder weights frozen. During evaluation, we report IoU and F1-scores on the spatio-temporal split. 

\begin{table}[t]
\centering
\caption{Parameter statistics for each GFM backbone with LoRA. 
Top: encoder only. Bottom: full encoder--decoder network.}
\label{tab:params_lora}
\resizebox{0.8\columnwidth}{!}{
\begin{tabular}{lccc}
\toprule
\textbf{Model} & \textbf{Total Params} & \textbf{Trainable (LoRA)} & \textbf{Percent} \\
\midrule
\multicolumn{4}{c}{\textbf{Encoder Only}} \\
\midrule
TerraMind   & 85{,}986{,}816  & 442{,}368  & 0.5145\% \\
DINOv3      & 86{,}112{,}000  & 442{,}368  & 0.5137\% \\
Prithvi-v2  & 306{,}245{,}632 & 3{,}145{,}728 & 1.0272\% \\
\midrule
\multicolumn{4}{c}{\textbf{Full Network}} \\
\midrule
TerraMind   & 100{,}937{,}410 & 15{,}392{,}962 & 15.2500\% \\
DINOv3      & 101{,}062{,}594 & 15{,}392{,}962 & 15.2311\% \\
Prithvi-v2  & 325{,}194{,}498 & 22{,}094{,}594 & 6.7943\% \\
\bottomrule
\end{tabular}
}
\end{table}

\section{Results and Discussion}

Table~\ref{tab:results} reports the performance of the three GFMs on the spatiotemporal split. 
Across all three foundation models, LoRA achieves the strongest performance, outperforming both full fine-tuning and decoder-only fine-tuning. Decoder-only fine-tuning consistently improves over full fine-tuning---by $+2.9$ IoU and $+1.9$ F1 for TerraMind, $+3.0$ IoU and $+2.0$ F1 for DINOv3, and $+1.6$ IoU and $+1.7$ F1 for Prithvi-v2---highlighting the benefit of freezing the backbone and updating only the task-specific decoder. LoRA further improves performance across all backbones, yielding an additional $+2.2$ IoU and $+1.4$ F1 for TerraMind, $+1.1$ IoU and $+0.7$ F1 for DINOv3, and a substantial $+6.8$ IoU and $+4.4$ F1 for Prithvi-v2. 
Among the three backbones, Prithvi-v2 exhibits the largest gains and the highest overall accuracy, suggesting that domain-specific pre-training yields stronger representations for wildfire-affected regions than generic vision models (DINOv3) or multimodal any-to-any models (TerraMind). The larger gains from LoRA on Prithvi-v2 are also consistent with its higher model capacity and its multi-temporal multi-resolution pretraining, whihc allows the low-rank updates to efficiently isolate fire-induced spectral changes. Under similar data and training settings, full fine-tuning of larger models is more prone to suboptimal convergence, making parameter-efficient adaptation more effective. Moreover, the consistent advantage of decoder-only tuning over full fine-tuning likely reflects a regularization effect, where updating all backbone parameters can lead to overfitting and degraded generalization. These results show that LoRA enables large EO encoders to better capture fire-related changes while keeping more than $99\%$ of backbone parameters frozen. As a result, GFMs can be efficiently adapted to burned-area mapping with minimal computational cost and strong cross-domain generalization, consistent with findings in prior work \cite{marti2025fine}.

\begin{table}
\centering
\small
\caption{Results of adaptation strategies on GFMs}
\label{tab:results}
\begin{tabular}{lccc}
\toprule
\textbf{Model} & \textbf{Adaptation} & \textbf{IoU} & \textbf{F1} \\
\midrule

\multirow{3}{*}{\textbf{TerraMind}}
& Full fine-tuning & 70.52 & 82.71 \\
& Decoder-only      & 73.39 & 84.65 \\
& LoRA              & \textbf{75.59} & \textbf{86.10} \\
\midrule

\multirow{3}{*}{\textbf{DINO-v3}}
& Full fine-tuning & 71.77 & 83.56 \\
& Decoder-only     & 74.72 & 85.53 \\
& LoRA             & \textbf{75.79} & \textbf{86.23} \\
\midrule

\multirow{3}{*}{\textbf{Prithvi-v2}}
& Full fine-tuning & 69.43 & 81.96 \\
& Decoder-only     & 71.98 & 83.71 \\
& LoRA             & \underline{\textbf{78.78}} & \underline{\textbf{88.13}} \\
\bottomrule
\end{tabular}
\end{table}

\begin{figure}[t]
    \centering

    \begin{minipage}{0.05\columnwidth}
    \scriptsize
        \rotatebox{90}{Input}
    \end{minipage}
    \begin{minipage}{0.2\columnwidth}
        \centering
        \scriptsize Pre-fire\\[3pt]
        \includegraphics[width=\linewidth]{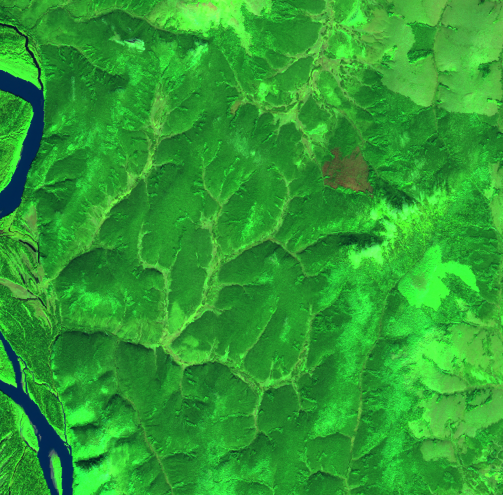}
    \end{minipage}
    \begin{minipage}{0.2\columnwidth}
        \centering
        \scriptsize Post-fire\\[3pt]
        \includegraphics[width=\linewidth]{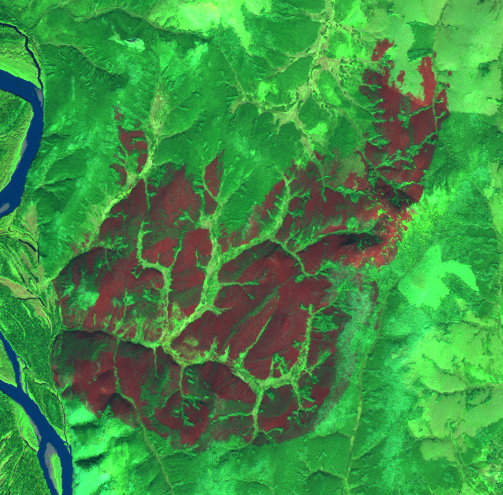}
    \end{minipage}
    \begin{minipage}{0.2\columnwidth}
        \centering
        \scriptsize Ground Truth\\[3pt]
        \includegraphics[width=\linewidth]{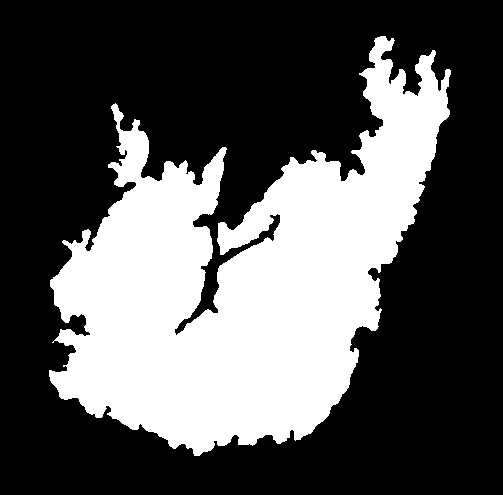}
    \end{minipage}

    \vspace{0.5cm}

    \begin{minipage}{0.05\columnwidth}
        \scriptsize
        \rotatebox{90}{TerraMind}
    \end{minipage}
    \begin{minipage}{0.2\columnwidth}
        \centering
        \scriptsize Full finetuning\\[3pt]
        \includegraphics[width=\linewidth]{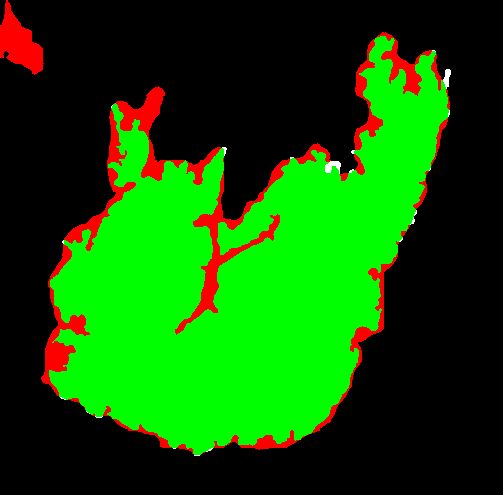}
    \end{minipage}
    \begin{minipage}{0.2\columnwidth}
        \centering
        \scriptsize Decoder-only\\[3pt]
        \includegraphics[width=\linewidth]{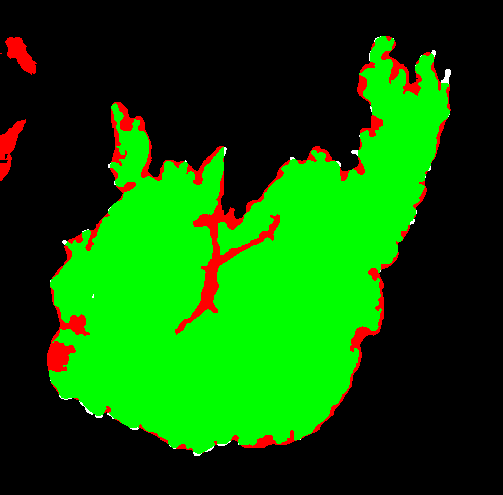}
    \end{minipage}
    \begin{minipage}{0.2\columnwidth}
        \centering
        \scriptsize LoRA\\[3pt]
        \includegraphics[width=\linewidth]{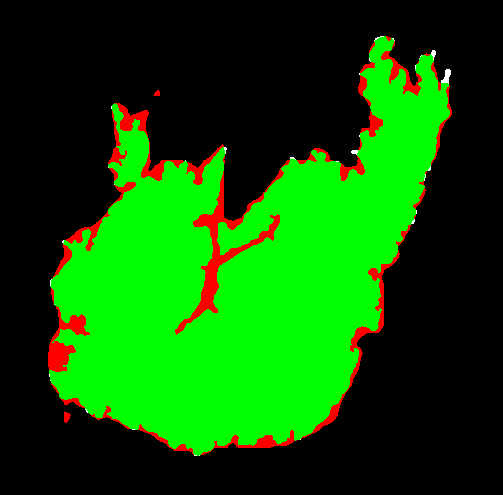}
    \end{minipage}
    \begin{minipage}{0.2\columnwidth}
        \centering
        \scriptsize Ground Truth\\[3pt]
        \includegraphics[width=\linewidth]{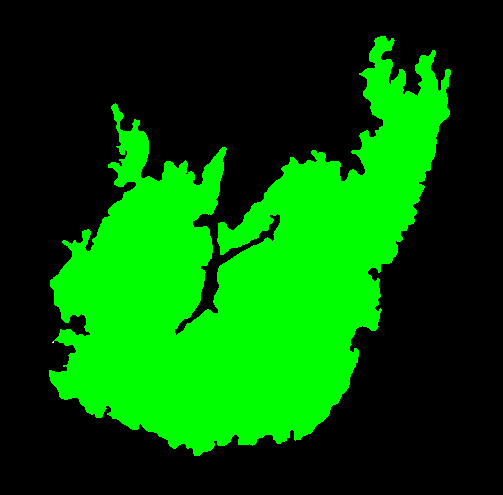}
    \end{minipage}
    
    \vspace{0.5cm}

    \begin{minipage}{0.05\columnwidth}
        \scriptsize
        \rotatebox{90}{Dinov3}
    \end{minipage}
    \begin{minipage}{0.2\columnwidth}
        \centering
        \scriptsize Full finetuning\\[3pt]
        \includegraphics[width=\linewidth]{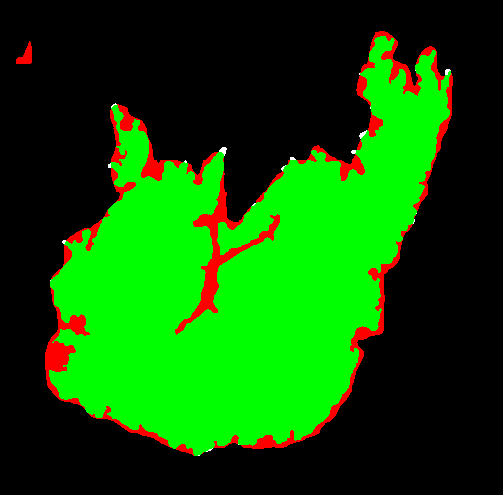}
    \end{minipage}
    \begin{minipage}{0.2\columnwidth}
        \centering
        \scriptsize Decoder-only\\[3pt]
        \includegraphics[width=\linewidth]{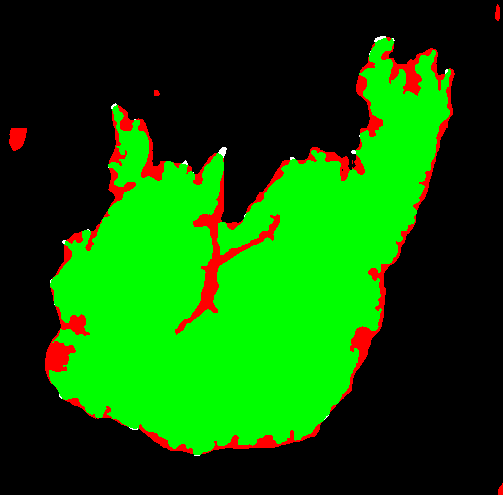}
    \end{minipage}
    \begin{minipage}{0.2\columnwidth}
        \centering
        \scriptsize LoRA\\[3pt]
        \includegraphics[width=\linewidth]{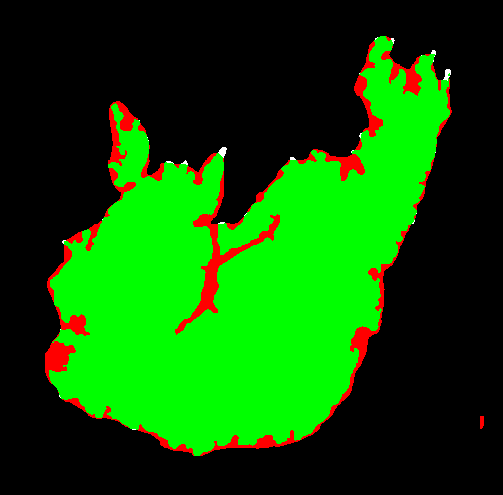}
    \end{minipage}
    \begin{minipage}{0.2\columnwidth}
        \centering
        \scriptsize Ground Truth\\[3pt]
        \includegraphics[width=\linewidth]{Figures/error_analysis/gt_AK6212515612820190711_medium_bw.png}
    \end{minipage}
    
    \vspace{0.5cm}

    \begin{minipage}{0.05\columnwidth}
        \scriptsize
        \rotatebox{90}{Prithvi-v2}
    \end{minipage}
    \begin{minipage}{0.2\columnwidth}
        \centering
        \scriptsize Full finetuning\\[3pt]
        \includegraphics[width=\linewidth]{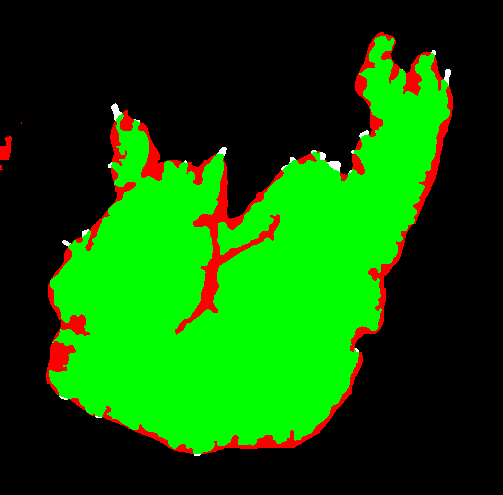}
    \end{minipage}
    \begin{minipage}{0.2\columnwidth}
        \centering
        \scriptsize Decoder-only\\[3pt]
        \includegraphics[width=\linewidth]{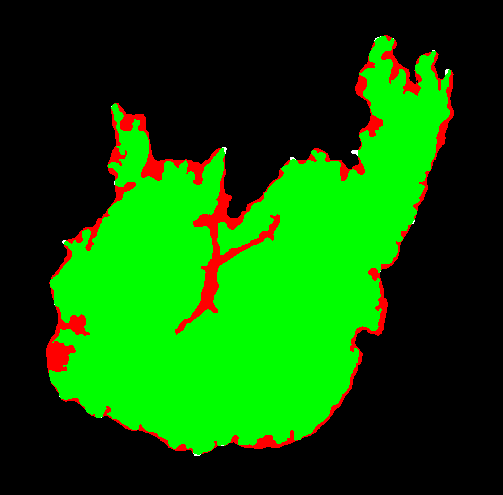}
    \end{minipage}
    \begin{minipage}{0.2\columnwidth}
        \centering
        \scriptsize LoRA\\[3pt]
        \includegraphics[width=\linewidth]{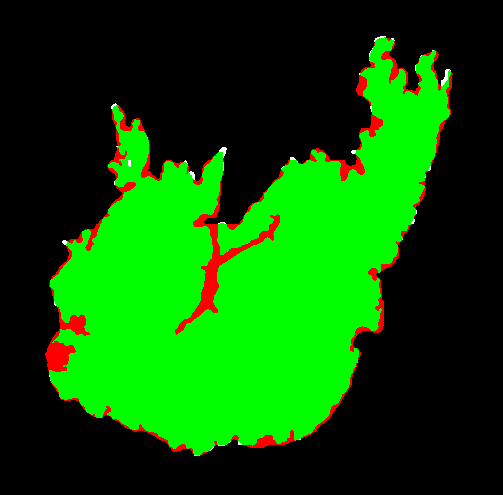}
    \end{minipage}
    \begin{minipage}{0.2\columnwidth}
        \centering
        \scriptsize Ground Truth\\[3pt]
        \includegraphics[width=\linewidth]{Figures/error_analysis/gt_AK6212515612820190711_medium_bw.png}
    \end{minipage}

    \caption{Qualitative full-fire burned-area predictions illustrating the effect of different adaptation strategies across three EO foundation models. For each backbone, we show results from full fine-tuning, decoder-only fine-tuning, and LoRA adaptation, alongside the ground-truth mask. Colors depicts the true positives in \textcolor{green}{green}, false positives in \textcolor{red}{red}, and false negatives in white.}
    \label{fig:fullfire_lora}
\end{figure}

For visual inspection, we plot full-fire burned area maps using the trained models and logit-averaging strategy since our models are trained on $128 \times 128$ pixel-patches. For each wildfire, we apply sliding-window inference (window size $128 \times 128$, stride $32$) with logit averaging over overlapping windows to reconstruct a full-scene prediction. 
While patch-based inference may introduce boundary artifacts or reduce spatial coherence, the use of overlapping windows with logit averaging mitigates these effects and improves prediction consistency across patch boundaries.
Figure~\ref{fig:fullfire_lora} shows qualitative examples for each backbone. We observe a clear progression from full fine-tuning to decoder-only to LoRA adaptation, where predictions become increasingly aligned with the ground-truth masks. In particular, LoRA reduces false positives along fire perimeters and suppresses and false negative detections within unburned regions.



Beyond parameter counts, the efficiency of LoRA translates into significant operational advantages for real-world deployment. By keeping the massive GFM backbone frozen, the required training computation is reduced, enabling faster convergence on standard hardware. In an operational pipeline, a single frozen foundation model can be deployed on a central server, while lightweight LoRA adapters can then be dynamically swapped in memory to process imagery from different geographic or seasonal fire regimes without needing to load entirely separate, fully fine-tuned models.

Finally, although our dataset is geographically focused on US and Canada, the environmental variance captured within this study serves as a strong proxy for global scalability. The successful transfer of the model across distinct biomes and years suggests that the learned representations are not strictly localized. Future applications to differing fire regimes, such as African or Australian forests, would require minimal target-domain data to train a region-specific LoRA adapter, ensuring the approach remains globally scalable.

\section{Conclusion}


This work demonstrates that parameter-efficient fine-tuning provides an effective and scalable way for adapting GFMs to large-scale burned-area mapping. Across more than 3,800 wildfire events, LoRA consistently outperforms both full and decoder-only fine-tuning while updating less than 1\% of encoder parameters. Prithvi-v2 shows the largest gains, highlighting the advantage of EO-specific pre-training, whereas TerraMind and DINOv3 achieve more modest but reliable improvements. LoRA also enhances robustness under spatiotemporal domain shift with minimal computational overhead, making it a practical choice for operational cross-domain wildfire monitoring. 
Future work will include integrating SAR imagery and expanding evaluation beyond North America.

\small
\bibliographystyle{IEEEtranN}
\bibliography{references}

@article{prithvi,
  title={Prithvi: Large-Scale Multimodal FMs for Earth Observation},
  author={Reichstein, Markus et al.},
  journal={NeurIPS},
  year={2023}
}

@article{terramind,
  title={TerraMind: Modality-Agnostic Geospatial Foundation Model},
  author={AWS AI Lab},
  journal={CVPR},
  year={2024}
}

@article{al2022burned,
  title={Burned area determination using Sentinel-2 satellite images and the impact of fire on the availability of soil nutrients in Syria.},
  author={Al-Hasn, Rukea and Almuhammad, Raed},
  year={2022}
}

@article{suwanprasit2024mapping,
  title={Mapping burned areas in Thailand using Sentinel-2 imagery and OBIA techniques},
  author={Suwanprasit, Chanida and Shahnawaz},
  journal={Scientific Reports},
  volume={14},
  number={1},
  pages={9609},
  year={2024},
  publisher={Nature Publishing Group UK London}
}

@article{knopp2020deep,
  title={A deep learning approach for burned area segmentation with Sentinel-2 data},
  author={Knopp, Lisa and Wieland, Marc and R{\"a}ttich, Michaela and Martinis, Sandro},
  journal={Remote Sensing},
  volume={12},
  number={15},
  pages={2422},
  year={2020},
  publisher={MDPI}
}

@article{brand2021semantic,
  title={Semantic segmentation of burned areas in satellite images using a U-net-based convolutional neural network},
  author={Brand, AK and Manandhar, A},
  journal={The International Archives of the Photogrammetry, Remote Sensing and Spatial Information Sciences},
  volume={43},
  pages={47--53},
  year={2021},
  publisher={Copernicus GmbH}
}

@article{sui2024biau,
  title={BiAU-Net: Wildfire burnt area mapping using bi-temporal Sentinel-2 imagery and U-Net with attention mechanism},
  author={Sui, Tang and Huang, Qunying and Wu, Mingda and Wu, Meiliu and Zhang, Zhou},
  journal={International Journal of Applied Earth Observation and Geoinformation},
  volume={132},
  pages={104034},
  year={2024},
  publisher={Elsevier}
}

@article{anand2025domain,
  title={Domain Adaptation and Fine-Tuning of a Deep Learning Segmentation Model of Small Agricultural Burn Area Detection Using High-Resolution Sentinel-2 Observations: A Case Study of Punjab, India},
  author={Anand, Anamika and Imasu, Ryoichi and Dhaka, Surendra K and Patra, Prabir K},
  journal={Remote Sensing},
  volume={17},
  number={6},
  pages={974},
  year={2025},
  publisher={MDPI}
}

@article{simeoni2025dinov3,
  title={Dinov3},
  author={Sim{\'e}oni, Oriane and Vo, Huy V and Seitzer, Maximilian and Baldassarre, Federico and Oquab, Maxime and Jose, Cijo and Khalidov, Vasil and Szafraniec, Marc and Yi, Seungeun and Ramamonjisoa, Micha{\"e}l and others},
  journal={arXiv preprint arXiv:2508.10104},
  year={2025}
}

@article{marsocci2024pangaea,
  title={Pangaea: A global and inclusive benchmark for geospatial foundation models},
  author={Marsocci, Valerio and Jia, Yuru and Bellier, Georges Le and Kerekes, David and Zeng, Liang and Hafner, Sebastian and Gerard, Sebastian and Brune, Eric and Yadav, Ritu and Shibli, Ali and others},
  journal={arXiv preprint arXiv:2412.04204},
  year={2024}
}

@article{simumba2025geo,
  title={Geo-bench-2: From performance to capability, rethinking evaluation in geospatial ai},
  author={Simumba, Naomi and Lehmann, Nils and Fraccaro, Paolo and Alemohammad, Hamed and De Mel, Geeth and Khan, Salman and Maskey, Manil and Longepe, Nicolas and Zhu, Xiao Xiang and Kerner, Hannah and others},
  journal={arXiv preprint arXiv:2511.15658},
  year={2025}
}

@article{han2024parameter,
  title={Parameter-efficient fine-tuning for large models: A comprehensive survey},
  author={Han, Zeyu and Gao, Chao and Liu, Jinyang and Zhang, Jeff and Zhang, Sai Qian},
  journal={arXiv preprint arXiv:2403.14608},
  year={2024}
}

@article{xin2024parameter,
  title={Parameter-efficient fine-tuning for pre-trained vision models: A survey},
  author={Xin, Yi and Luo, Siqi and Zhou, Haodi and Du, Junlong and Liu, Xiaohong and Fan, Yue and Li, Qing and Du, Yuntao},
  journal={arXiv e-prints},
  pages={arXiv--2402},
  year={2024}
}

@article{hu2022lora,
  title={Lora: Low-rank adaptation of large language models.},
  author={Hu, Edward J and Shen, Yelong and Wallis, Phillip and Allen-Zhu, Zeyuan and Li, Yuanzhi and Wang, Shean and Wang, Lu and Chen, Weizhu and others},
  journal={ICLR},
  volume={1},
  number={2},
  pages={3},
  year={2022}
}

@inproceedings{marti2025fine,
  title={Fine-tune smarter, not harder: Parameter-efficient fine-tuning for geospatial foundation models},
  author={Marti Escofet, Francesc and Blumenstiel, Benedikt and Scheibenreif, Linus and Fraccaro, Paolo and Schindler, Konrad},
  booktitle={Joint European Conference on Machine Learning and Knowledge Discovery in Databases},
  pages={516--532},
  year={2025},
  organization={Springer}
}

@inproceedings{xiao2018unified,
  title={Unified perceptual parsing for scene understanding},
  author={Xiao, Tete and Liu, Yingcheng and Zhou, Bolei and Jiang, Yuning and Sun, Jian},
  booktitle={Proceedings of the European conference on computer vision (ECCV)},
  pages={418--434},
  year={2018}
}

@article{mtbs,
  title={A project for monitoring trends in burn severity},
  author={Eidenshink, Jeff and Schwind, Brian and Brewer, Ken and Zhu, Zhi-Liang and Quayle, Brad and Howard, Stephen},
  journal={Fire ecology},
  volume={3},
  number={1},
  pages={3--21},
  year={2007},
  publisher={Springer}
}

@misc{nbac,
  author = {{Canadian Forest Service}},
  title  = {National Burned Area Composite ({NBAC}) --- Annual Burned Area Polygons},
  year   = {2023},
  note   = {Government of Canada metadata catalogue.}
}

\end{document}